# An approach of IR-Based short-range correspondence systems for swarm robot balanced requisitions and communications

Rafiqul Islam ,Iftekharul Mobin , Md. Nazmus Shakib, Md Matiur Rahman

Department of Computer Science and Engineering

University of Liberal Arts Bangladesh (ULAB)

House No. 56, Road No. 4/A, Satmasjid Road, Dhaka 1209, Bangladesh

rafiqul713@gmail.com, iftekharul.mobin@ulab.edu.bd, nazmus.shakib.cse@gmail.com, matiur.rahman.cse@ulab.edu.bd

*Abstract:*

This paper exhibits a short-run correspondence method appropriate for swarm versatile robots application. Infrared is utilized for transmitting and accepting information and obstruction location. The infrared correspondence code based swarm signaling is utilized for an independent versatile robot communication system in this research. A code based signaling system is developed for transmitting information between different entities of robot. The reflected infrared sign is additionally utilized for separation estimation for obstruction evasion. Investigation of robot demonstrates the possibility of utilizing infrared signs to get a solid nearby correspondence between swarm portable robots. This paper exhibits a basic decentralized control for swarm of self-collecting robots. Every robot in the code based swarm signaling is completely self-governing and controlled utilizing a conduct based methodology with just infrared-based nearby detecting and correspondences. The viability of the methodology has been checked with simulation, for a set of swarm robots.

## 1. Introduction:

Nowadays swarm robots have become interesting fields. In this work we developed an IR-based swarm robot communication. IR has also become another medium of communication. Here we used IR because this allows the robot to minimize the power consumption. It needs very little amount of energy to operate. Each robot can detect another robot (homologous) and communicate with each other. Each robot can also avoid obstacle. After observing the environment each robot can get reliable information about other robot and obstacle and their position in the environment.  Swarm robots are a large number of collective agents and they work cooperatively to solve different task. One robot can recognize the other homologous robot through emitted IR radiation. Each robot can take decision based on observing the environment and analyzing it. They can communicate themselves according to their needs and to full-fill their purpose. The robot that are sent out alone or in non-cooperative pairs there is a high risk of failure. In this case swarm robot can be sent out on a mission with a high level of confidence.

In swarm robotics research we experienced numerous issues, for example, coordination rules aggregate basic swarm leadership [1] with scientific and algorithmic approaches [1-5]. All the more for the most part, these issues are identified with various fields of AI [5] and outlining of swarm robotics [6], [7] e.g. through exemplification  in any case every one of these instruments work just when robots are associating [8].

There are a few methods for how the robots can cooperate: robots watch environment and the conduct of different robots, physical association, backhanded cooperation trough environment or they impart. Since the micro robots are confined in detecting and calculation, this method for associating group of swarms will be appropriate [4]. There are works on e.g. acknowledgment of robots by transmitted IR-radiation [8], shading observation then again utilizing impacts as connections among robots.

Complex psychological and behavioral systems are incomprehensible without correspondence. Creating correspondence systems for certain groups of swarm [9] experiences couple of issues for implementation. In the first place of all, robots have just constrained correspondence sweep. This permits maintaining a strategic distance from the issue of correspondence flood swarms (100+ robots), be that as it may opens the issue of proliferating the applicable data over the swarm. This data concerns e.g. vitality assets, behavioral objectives, threats and so on. Robots are confined in equipment for utilizing calculations and conventions known in the area of dispersed frameworks [10]. Thusly new ideas and new conventions ought to be produced for the swarm correspondence. Not just programming conventions, additionally correspondence equipment ought to be adjusted to the need of expansive scale swarms. This essentially concerns a multi-channel hardware for omni directional swarm neighborhood correspondence, utilizing of low-level signs, streamlining of the transmitted vitality and arrangement of directing issue [11]. We say that exclusive a great interchange between equipment, programming and robots conduct permits a solid data move in a swarm mechanical autonomy frameworks are described by decentralized control, constrained correspondence between robots, use of neighborhood data, and development of worldwide conduct. Such frameworks have demonstrated their potential for adaptability and power [1]–[3]. In any case, existing swarm mechanical autonomy frameworks are all things considered still constrained to showing basic verification of idea practices under lab conditions. Swarm robots must grasp heterogeneity that they ever need to approach the intricacy required of frameworks. To date, swarm mechanical technology frameworks have only contained physically, behaviorally undifferentiated specialists[13]. This configuration choice has its roots in code based swarm signaling models of self-sorting out characteristic frameworks. These models serve as motivation for swarm mechanical technology code based swarm signaling architects, yet are regularly exceptionally conceptual rearrangements of characteristic frameworks. Chosen elements of the frameworks under study are appeared to rise up out of the communications of indistinguishable code based swarm signaling parts, disregarding the heterogeneities (physical, spatial, practical, and instructive) that one can discover in any common framework [14] [15]. The field of swarm mechanical technology presently needs techniques and instruments with which to study and influence the heterogeneity that is available in characteristic frameworks. To cure this lack, we propose a code based infrared IR communication system, a creative swarm mechanical technology code based swarm signaling made out of three distinctive robot sorts with correlative abilities. So that the root can operate as self-governing robots spent significant time in proceeding onward both even and uneven landscapes, fit for self-collecting and of transporting objects; self-sufficient fit for climbing some vertical surfaces and controlling little protests. Also can be appended for investigating the earth from an advantaged position.

## 2. Literature Review:

A group of mobile robots builds one pile from some randomly distribute object[15],[16]. The mobile robots group coordinates with indirect communication through sensing and modification of the local environment which determine the agent behavior from Local Action to Global Task. A multiple robots system performs their task without centralize control or explicit communication. Mobile robots able to achieving simple collective task and verify result by simulation. The result suggests cooperative task requiring collective behavior without explicit communication [17]. Jasmin robot allows huge scale swarm system investigate artificial self organization, control large robotic group [18]. For later development from 2007-2010 revise these idea swarm artificial where include evolutionary robotics, online embodied evolving, system capable structure modification and general field of adaptive system [19]. This facilitate more embodied cognition on swarm and collective robots.- IR-based Communication and Perception in Micro-robotic swarm. Swarm mobile robots application used short range communication technique [18], [19]. In swarm robotics infrared is used to transmitting and receiving data packet and obstacle detection. This swarm robots application use PCM (Pulse-Code-Modulation) digital scheme for transmitting data [20]. Communication with mobile robots by infrared signal which is depends on robots behavior [21]. For Collective combined behavior use custom range and bearing system base on cascaded filtering technology implemented by infrared frequency communication. Infrared Range and Bearing system for Collective Robots [21]. For observation mobile combined behavior use 360-degree observation capability which identification of neighboring robots from object and their position. For observation use DRIr ( Dual Rotating Infrared) sensor which is based on algorithm. Sensor provides real time location such as Dual Rotating Infrared. Similar robotics application of combined behavior is implemented by Melhuish. For our application we used infrared for communication [20], [21].

## 3. Equipment uses: Most of the equipment used in this work is related to the

### 3.1 Ardunio Uno:
Ardunio Uno is a micro-controller board based on the ATmega328P. It has 14 digital input/output pins (of which 6 can be used as PWM outputs), 6 analog inputs, a 16 MHz quartz crystal, a USB connection, a power jack, an ICSP header and a reset button. It contains everything needed to support the microcontroller; simply connect it to a computer with a USB cable or power it with an AC-to-DC adapter or battery to get started.

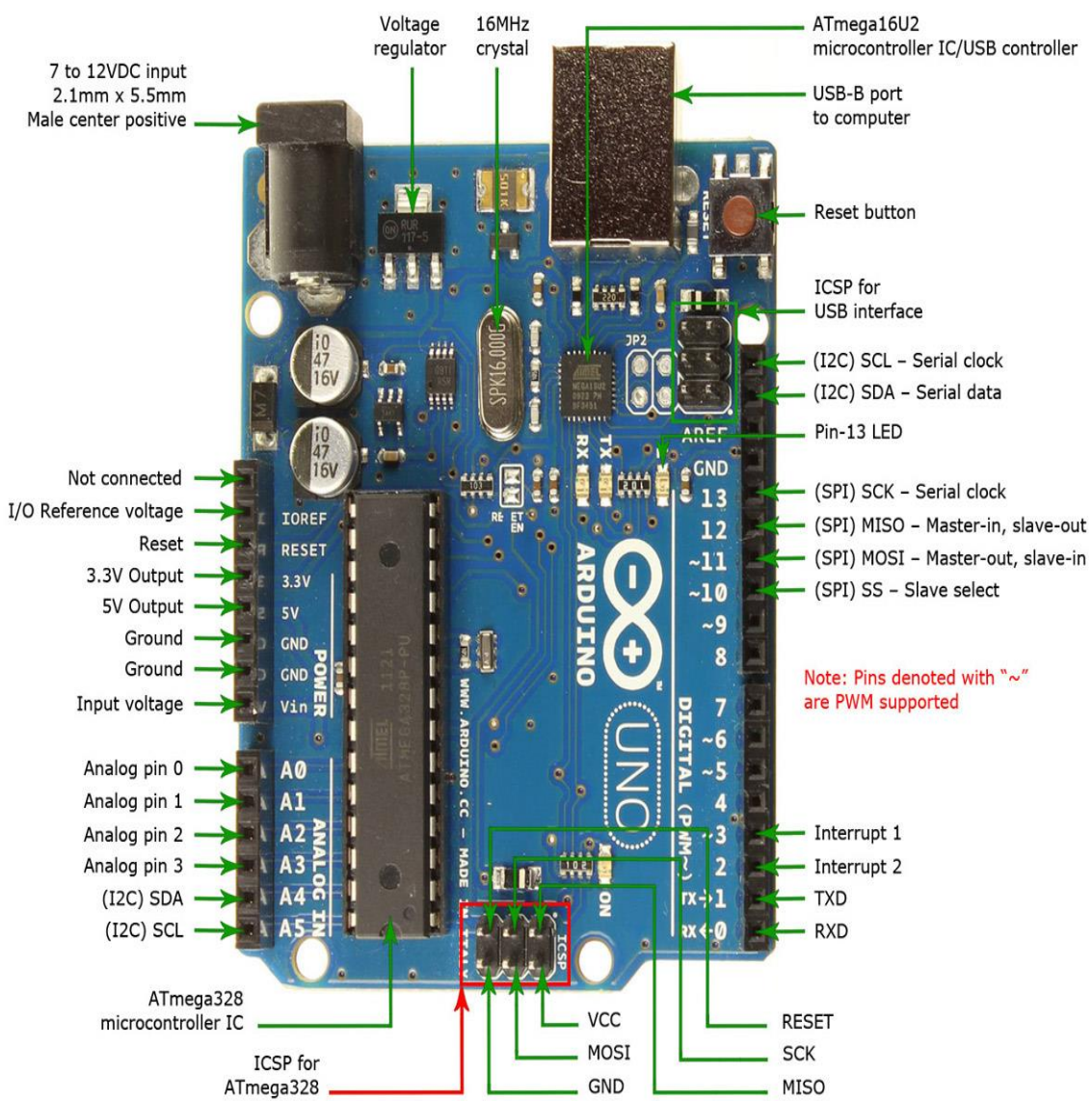

Fig 1.1: Ardunio Unu [23]

## 3.2 IR communication:

IR, or infrared, communication is a common, inexpensive, and easy to use wireless communication technology. IR light is very similar to visible light, except that it has a slightly longer wavelength. This means IR is undetectable to the human eye - perfect for wireless communication.

## 3.3 IR transmitter:

The IR transmitter consists of the LED that emits the IR(Infra Red) radiation. This is received by the photo diode, which acts as IR receiver at the receiving end.

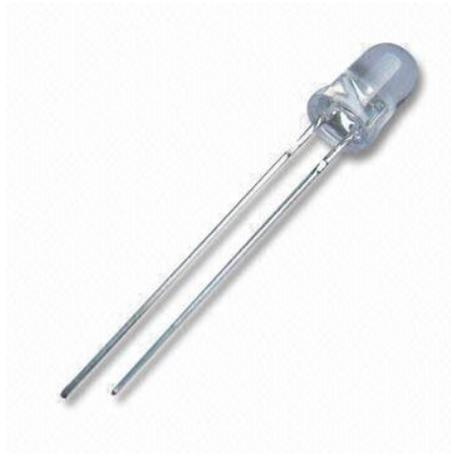

Fig 1.2: IR Transmitter[24]

## 3.4 IR receiver:

An infrared receiver, or IR receiver, is hardware that sends information from an infrared remote control to another device by receiving and decoding signals. In general, the receiver outputs a code to uniquely identify the infrared signal that it receives.

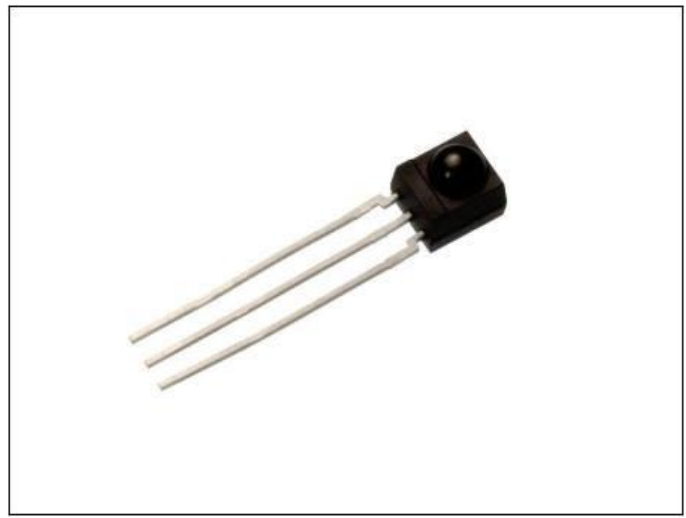

Fig 1.3: IR receiver[25]

## 4. Flow Chart

# Robot Robot Communication With IR

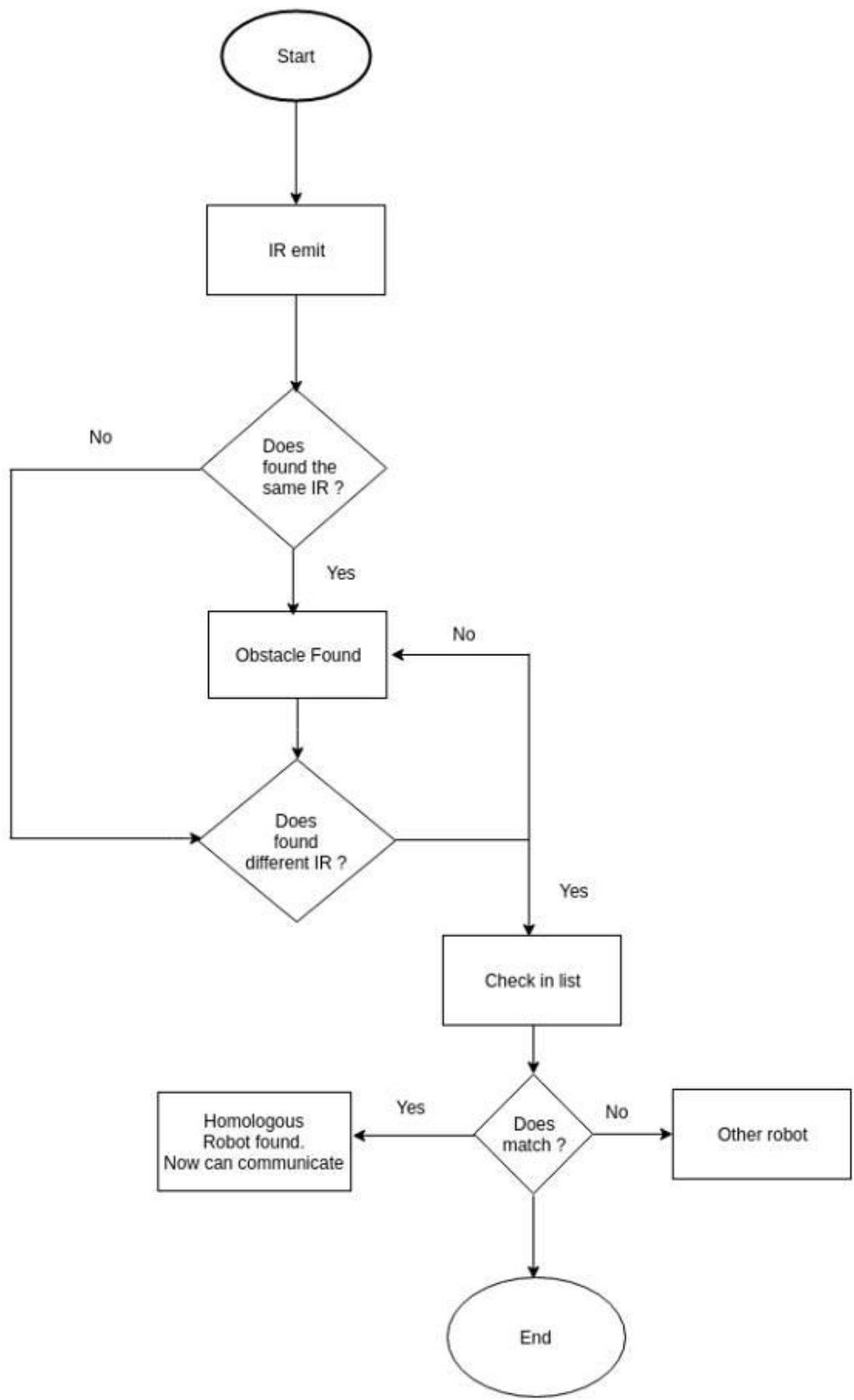

Flowchart Description:

After starting swarm robot, at first each swarm robot emits IR radiation spontaneously through IR transmitter. IR receiver will receive the IR signal. If it receives the same IR signal which it sent then it will decide there are obstacle in-front of it. If it receive different IR signal then check the list. If the list contains this IR signal then it will take decide there are Homologous Robot in-front of it. Now it is ready to communicate with them. If list does not contain the received IR then it will take decision there are other IR transmitting source here.

## 5. Circuit Block Diagram:

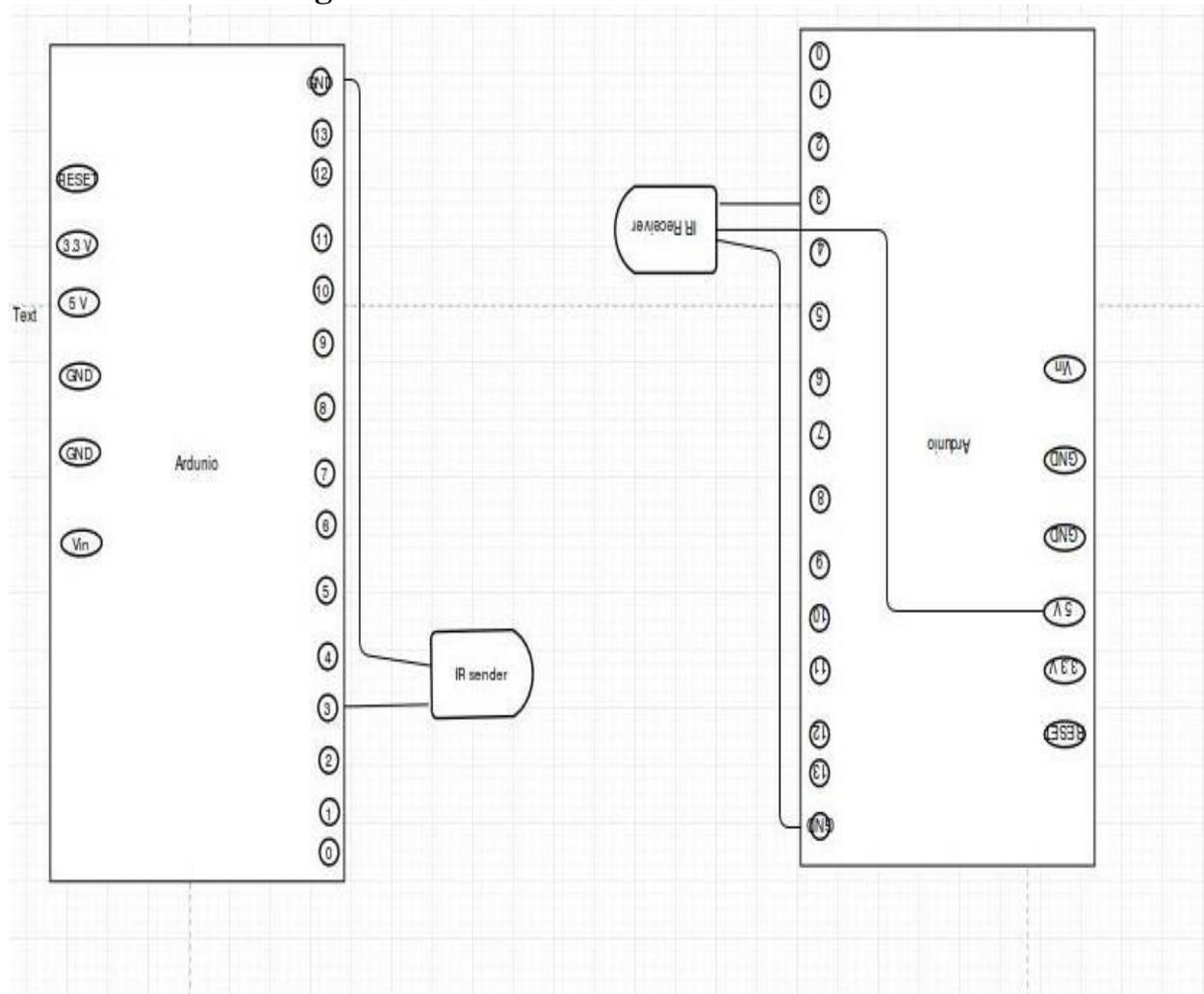

Fig 1.4: **Block Diagram**

In this block diagram there are two Ardunio are used for describing IR communication between swarm robots. Each Ardunio contain one IR transmitter/sender and one IR receiver.

Left side Ardunio, IR sender's one pin is connected to the ground pin of Ardunio and another one is connected to Ardunio's Digital Pin 3.
Right side Ardunio, IR receiver's one pin is connected to 5 volt of Ardunio another one is connected to the ground of Ardunio and other one pin is connected to the Ardunio's digital pin 3.

## 6. Pseudo Code:

```
1. Initialize array to store Identification number of Homogeneous robot.
2. Store Identification number in IR Database
3. while(TRUE)
        3.1. Emit IR Code
        3.2 if receive the same IR Code
                3.2.1. Obstacle Found
        3.3. else if receive different IR Code
                3.3.1.1 Check the receive data in IR Database
                3.3.1.2 if it match with IR Database entry
                        3.3.1.2.1 The Robot is Familiar
                        3.3.1.2.2 now can Communicate
                3.1.1.3 else
                        3.3.3.1Other Robots or IR emitted from unknown sources
        3.4. else There is not any obstacle in-front of the robot
```

At first initialize the total number of homologous swarm robot/agent which will operate in the environment. Each robot will be given a unique identification number which distinguish them from others. The condition of while loop is true. It will execute infinitive time. The scope of while loop will execute infinitive time. Here it will emit IR code. Then IR receiver will receive the IR code. If it receive the same IR code then obstacle will found. If it receives different IR then check the IR code in list which is initialize first time. If received IR code matched with the list then the robot is familiar. Now it is ready to communicate with it. If it receives other IR code then it will decide other IR sources are existing in the environment. On the other hand if it does not receive any IR code then it will decide there are no obstacle in-front of it and no homogeneous robot in the environment.

## 7. Methodologies and Protocol:

Here, we develop a protocol for communication. Each robot sends a request to another robot for communication. If another robot is homologous robot then it accepts the request and they communicate each other by sending IR signal. Synchronize illustration: This protocol demonstrate how a gathering of robots can convey about headings where they are indicating encompassing neighbors and wind up adjusted towards the same course. In a complete circle, the robots attempt to:

i) Send data: Because of the quantity of robots, every robot will first send his own ID, and after that his own heading identified with the robots from which it has got a message.

ii) Listen for a few transmissions and unravel the messages. In the event that a message is an ID, it will store the heading where it comes from. On the off chance that the message is an introduction tended to it, it will change the introduction, as indicated by this new data.

### 7.1 Investigation:

Wrong messages got, or messages not got. Robots transmit the same infrareds. So when a few robots attempt to send in the same time, most likely get just wrong messages. IR interchanges are marginally touchy to light conditions. On the off chance that need to draw near to 0 messages lost, attempt to work with controlled light, discharging couple of infrareds, and without motions. The infrared on utilized is set to a high need which may supersede others. To adapt these issues of timing, we have executed a period counter inside Arduino code. The algorithm provides exact timings.

### 7.2 IR emission process:

The first phase is IR transmission. Each robot is equipped with an IR transmitter and it will continuously emit IR. Here it sends particular data through IR transmitter. IR transmitter is use for two purposes. First reason, Detecting Obstacle; second reason, Communication purpose.

### 7.3 IR receiving process:

IR receiver is used for receiving IR. Every robot has one IR receiver. Each robot receives IR data then matches it with its database. If it receives the same IR data that it sends in this case there is an obstacle in front of this robot. That's why it receives the same data which is reflected to the obstacle. If robot receives different data then it compares the data with its database. If it matches in this case another homologous robot is found. Now it can communicate with that robot. Otherwise it does not interest to communicate.

## 8. Conclusion

In this work we exhibited swarm robotics equipment set up and programming issues concerning IR-based communication and correspondence. We have demonstrated that few challenges made by constrained equipment capacities can be effectively overcome. The purpose of the exhibited investigations is identified with exemplification. This ought to make another motivation towards reconfigurable and developmental methodologies for swarm robot communication system. To this end, it is important to create apparatuses and procedures that empower the utilization of swarm robot communication for heterogeneous frameworks. We recognized significant issues and challenges, specifically highlighting the trouble of conveying the firmly incorporated automated equipment to empower physical and behavioral communication between various robot sorts.

## 9. Future Plan

Our future plan is to add more functionality with each swarm robot. To add GPS and IoT ( Internet of Things)  device with them so that they can be track anywhere in the world. To add mechanical hand with them so that they can do any work more efficient manner. Our future plan is to apply Artificial Neural Network in swarm robotics. But it will be a very challenging task. This swarm robot can be sent any environment they can adapt themselves. They can learn from their experience and become completely autonomous.

## REFERENCES:

[1]  Fong, T., Nourbakhsh, I., & Dautenhahn, K. (2003). A survey of socially interactive robots. *Robotics and*

*autonomous systems*, *42*(3), 143-166.

[2] Navarro, Iñaki, and Fernando Matía. "A survey of collective movement of mobile robots." *International Journal of Advanced Robotic Systems* 10.73 (2013).

[3] Willke, Theodore L., Patcharinee Tientrakool, and Nicholas F. Maxemchuk. "A survey of inter-vehicle communication protocols and their applications."*Communications Surveys & Tutorials, IEEE* 11.2 (2009): 3-20.

[4] Xiong, Naixue, et al. "A survey on decentralized flocking schemes for a set of autonomous mobile robots." *Journal of Communications* 5.1 (2010): 31-38.

[5] Portugal, David, and Rui Rocha. "A survey on multi-robot patrolling algorithms." *Technological innovation for sustainability*. Springer Berlin Heidelberg, 2011. 139-146.

[6] Cao, Y. Uny, Alex S. Fukunaga, and Andrew Kahng. "Cooperative mobile robotics: Antecedents and directions." *Autonomous robots* 4.1 (1997): 7-27.

[7] Tuci, Elio, et al. "Evolving homogeneous neurocontrollers for a group of heterogeneous robots: Coordinated motion, cooperation, and acoustic communication." *Artificial Life* 14.2 (2008): 157-178.

[8] Yan, Zhi, Nicolas Jouandeau, and Arab Ali Cherif. "A survey and analysis of multi-robot coordination." *International Journal of Advanced Robotic Systems*10 (2013).

[9] Berman, Spring, et al. "Bio-inspired group behaviors for the deployment of a swarm of robots to multiple destinations." *Robotics and Automation, 2007 IEEE International Conference on*. IEEE, 2007.

[10] Dutta, Indranil, A. D. Bogobowicz, and J. J. Gu. "Collective robotics-a survey of control and communication techniques." *Intelligent Mechatronics and Automation, 2004. Proceedings. 2004 International Conference on*. IEEE, 2004.

[11] Yang, Erfu, and Dongbing Gu. *Multiagent reinforcement learning for multi-robot systems: A survey*. tech. rep, 2004.

[12] Agmon, Noa, et al. "The giving tree: constructing trees for efficient offline and online multi-robot coverage." *Annals of Mathematics and Artificial Intelligence* 52.2-4 (2008): 143-168.

[13] Arai, Tamio, Enrico Pagello, and Lynne E. Parker. "Editorial: Advances in multi-robot systems." *IEEE Transactions on robotics and automation* 18.5 (2002): 655-661.

[14] Rooker, Martijn N., and Andreas Birk. "Multi-robot exploration under the constraints of wireless networking." *Control Engineering Practice* 15.4 (2007): 435-445.

[15] Penders, Jacques, et al. "Guardians: a swarm of autonomous robots for emergencies." *Proceedings of the 20th International Joint Conference on Artificial Intelligence (IJCAI'07) Workshop on Multirobotic Systems for Societal Applications*. 2007.

[16] Szu, Harold, et al. "Collective and distributive swarm intelligence: evolutional biological survey." *International Congress Series*. Vol. 1269. Elsevier, 2004.

[17] Mosteo, Alejandro R., and Luis Montano. "A survey of multi-robot task allocation." *Instituto de Investigación en Ingeniería de Aragón, University of Zaragoza, Zaragoza, Spain, Technical Report No. AMI-009-10-TEC* (2010).

[18] Potop-Butucaru, Maria, Michel Raynal, and Sébastien Tixeuil. "Distributed computing with mobile robots: an introductory survey." 2011 International C*onference on Network-Based Information Systems*. IEEE, 2011.

[19] Ota, Jun. "Multi-agent robot systems as distributed autonomous systems."*Advanced engineering informatics* 20.1 (2006): 59-70.

[20] Hazon, Noam, Fabrizio Mieli, and Gal A. Kaminka. "Towards robust on-line multi-robot coverage." *Robotics and Automation, 2006. ICRA 2006. Proceedings 2006 IEEE International Conference on*. IEEE, 2006.

[21] Pugh, Jim, and Alcherio Martinoli. "Multi-robot learning with particle swarm optimization." *Proceedings of the fifth international joint conference on Autonomous agents and multiagent systems*. ACM, 2006.

[22] "Swarmrobot | Open-Source Micro-Robotic Project". Swarmrobot.org. N.p., 2016. Web. 1 June 2016.

[23] www.xmhlec.manufacturer.globalsources.com, accessed: June, 2016

[24] "Solarbotics". Solarbotics.com. N.p., 2016. Web. 1 June 2016.

[25] Díaz-Boladeras, Marta, et al. "Evaluating group-robot interaction in crowded public spaces: A week-long exploratory study in the wild with a humanoid robot guiding visitors through a science museum." *International Journal of Humanoid Robotics* 12.04 (2015): 1550022.

[26] Heinerman, Jacqueline, Dexter Drupsteen, and A. E. Eiben. "Three-fold adaptivity in groups of robots: The effect of social learning." *Proceedings of the 2015 on Genetic and Evolutionary Computation Conference*. ACM, 2015.

[27] Singh, D. Narendhar, and G. Ashwini. "Swarm Robots for Environmental Monitoring and Surveillance."

[28] Sathiyanarayanan, Mithileysh. "Robots for Military Purpose."

[29] Hati, Monalisa. "Swarm Robotics: A Technological Advancement for Human-Swarm Interaction in Recent Era from Swarm-Intelligence Concept

**Biography**

Rafiqul Islam is a student in computer science and Engineering department of University of Liberal Arts Bangladesh(ULAB). He is an undergraduate student. His research interests are Embedded system, Artificial Intelligence,Artificial Neural Network, Linked Data,Digital Image Processing. He likes ACM programming.

Iftekharul Mobin is an Assistant Professor in computer Science and Engineering department of University of Liberal Arts Bangladesh (ULAB). He obtained his doctor's degree from the School of Electronic Engineering of Queen Mary, University of London in 2014. He completed his B.Sc in Computer Science and Information Technology (CIT) from Islamic University of Technology (IUT), Gazipur, Bangladesh in 2008. After his PhD he worked in a Telecommunication service provider company Natterbox Ltd in London as a researcher and developer in company R&D. His current research interests are artificial Intelligence, Embedded systems, Wireless Ad hoc Networks, Smart devices Robotics and sensor based technologies. He is the founder of wireless sensor and Robotics lab in ULAB. Dr. Mobin is an author and co-author of many International Journals and conference papers. He is currently professional member and student branch coordinator of IEEE in ULAB.

Md. Nazmus Shakib is a student in computer science and Engineering department of University of Liberal Arts Bangladesh(ULAB). He is an undergraduate student . He passed HSC from Naogaon Govt: College. His research interests are Embedded system.

Md Matiur Rahman is a student in computer science and engineering department of University of Liberal Arts Bangladesh(ULAB). He is an undergraduate student. His research topics is Digital Image Processing, Artificial Intelligence , Wireless Communication.